\documentclass[11pt]{article}
\usepackage{eamt24}
\usepackage{times}
\usepackage{url}
\usepackage{latexsym}
\usepackage[small,bf]{caption} 
\usepackage{hyperref}
\usepackage{booktabs}
\usepackage[dvipsnames]{xcolor}
\usepackage{amsmath}
\usepackage{csquotes}
\usepackage{multirow}
\usepackage{colortbl}
\usepackage{pifont}
\usepackage{subfig}
\usepackage{ulem}
\usepackage{yhmath}
\usepackage{enumitem}

\newcommand{\AfA}{\textsuperscript{\begin{minipage}[c]{2.5mm}
\includegraphics[width=\linewidth]{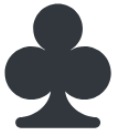}
\end{minipage}}}
\newcommand{\AfB}{\textsuperscript{\begin{minipage}[c]{2.5mm}
\includegraphics[width=\linewidth]{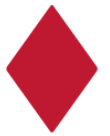}
\end{minipage}}}

\newcommand{\subs}{\textsc{ZOOSubs}}
\newcommand{\cmark}{\textcolor{OliveGreen}{\ding{51}}}
\newcommand{\xmark}{\textcolor{BrickRed}{\ding{55}}}
\newcommand{\brick}[1]{\textcolor{BrickRed}{#1}}
\newcommand{\olive}[1]{\textcolor{OliveGreen}{#1}}

\newcommand{\Reff}{\cellcolor{Peach!30!white}}
\newcommand{\BaseN}{\cellcolor{OliveGreen!40!white}}
\newcommand{\Goog}{\cellcolor{SkyBlue!30!white}}
\newcommand{\MTQ}{\cellcolor{Fuchsia!30!white}}
\title{A Case Study on Contextual Machine Translation\\in a Professional Scenario of Subtitling}

\author{\AfA \AfB Sebastian Vincent,
   	\AfB \textbf{Charlotte Prescott}
   	\AfB \textbf{Chris Bayliss},\\
   	\AfB \textbf{Chris Oakley},
   	\AfA \textbf{Carolina Scarton}
   	\\[5pt]
   	\AfA Department of Computer Science, University of Sheffield, UK\\
   	\AfB ZOO Digital Group PLC, UK\\}

\date{}

\begin{document}
\maketitle
\begin{abstract}
Incorporating extra-textual context such as film metadata into the machine translation (MT) pipeline can enhance translation quality, as indicated by automatic evaluation in recent work. However, the positive impact of such systems in industry remains unproven. We report on an industrial case study carried out to investigate the benefit of MT in a professional scenario of translating TV subtitles with a focus on how leveraging extra-textual context impacts post-editing. We found that post-editors marked significantly fewer context-related errors when correcting the outputs of \textsc{MTCue}, the context-aware model, as opposed to non-contextual models. We also present the results of a survey of the employed post-editors, which highlights contextual inadequacy as a significant gap consistently observed in MT. Our findings strengthen the motivation for further work within fully contextual MT.
\end{abstract}

\section{Introduction}
As an innovation-driven company offering dubbing and subtitling services, ZOO Digital is dedicated to exploring assistive technologies to streamline our workflows. Machine translation in particular is a promising tool for improving the efficiency of the (currently fully manual) translation of the transcribed video content during interlingual subtitling. Our domain is characterised by specific challenges, both linguistic (preservation of style and function in dialogue) and practical (keeping within subtitle constraints, such as visual properties and considerations for the viewers' reading speed). We report on a case study where translation from scratch was replaced with post-editing machine translations of the source text. While such a formulation is far from new -- MT has been consistently demonstrated to help reduce effort in the subtitling domain \cite{c-m-de-sousa-etal-2011-assessing,jie2019} -- previous studies have relied on off-the-shelf general-purpose neural machine translation (NMT) engines like Google Translate\footnote{\url{https://translate.google.com/}}. Our work investigates two additional systems: \textsc{Base-NMT}, a specialised engine trained on our data, as well its contextual version based on the \textsc{MTCue} architecture~\cite{vincent-etal-2023-mtcue}, whose training involves observing a vast range of metadata and document-level information. 

The study was carried out with the assistance of translation and post-editing professionals. Hereinafter we refer as \textit{post-editors (PEs)} to those who were tasked with post-editing work, and as \textit{translators (HTs)} to those who were tasked with translation from scratch (FST). The campaign took place in a full-context multi-modal environment where the professionals had access to the video material and were able to directly jump to the segment corresponding to the utterance they were reviewing, as well as see the preceding and succeeding segments. A total of eight PEs were employed, four for English-to-German (\textsc{en-de}) and four for English-to-French (\textsc{en-fr}) translation, and four HTs, two per language pair. We measured the effort it took to post-edit or translate the TV series content and the number of specific translation errors observed by the PEs. Our findings highlight the necessity of tailoring MT engines to the target domain and motivate further work within leveraging contextual systems in dialogue translation.

\section{Related Work} \label{eamt24:rw}
Over the last few years, subtitle translation has been given a volume of attention:~\newcite{c-m-de-sousa-etal-2011-assessing},~\newcite{koponen-etal-2020-mt} and~\newcite{jie2019} observe that post-editing the outputs of an NMT system is a promising alternative to translation \textit{ex novo}, reducing the temporal, technical and cognitive effort of both novice and professional translators and subtitlers. A survey among professional subtitlers detailed by~\newcite{karakanta-etal-2022-post}, finds that professionals have a positive outlook on incorporating automatic components (such as MT) into their workflow, as they offer starting templates, reduce effort and can provide useful suggestions. However, some challenges in the automatic translation of subtitles remain unsolved~\cite{gupta-2019-subtitles,karakanta-etal-2022-post}, including the adherence to subtitle block limitations, which often necessitates shorter and paraphrased translations; lexical consistency, which involves translating the same terms across the text, as well as using vocabulary that maintains the cohesion and coherence of the text, aligns with the surrounding video or textual content, and conforms to standard language or industry conventions; lexical errors such as the translation of idioms and figurative language, and context-related inconsistencies. Context-related errors in particular have been pointed out as the culprit in many works in MT that leveraged the OpenSubtitles corpus~\cite{lison-etal-2018-opensubtitles2018}, a dataset of user-submitted subtitles and their translations. Leveraging document-level information~\cite{tiedemann-scherrer-2017-neural,Bawden2018}, speaker's and interlocutor's gender identity~\cite{vincent-etal-2022-controlling-extra} and explicit extra-textual information~\cite{vincent-etal-2023-mtcue} has been found particularly useful in addressing this challenge. Context is also useful during the manual post-editing procedure: ~\newcite{jie2019} show that such a setup decreases the cognitive load of student translators compared to a text-only scenario, suggesting as an explanation the dual coding theory, according to which the interactions between the verbal and non-verbal information enhances the translators' understanding of the material.

This work employs \textsc{MTCue}~\cite{vincent-etal-2023-mtcue}, a multi-encoder Transformer designed for contextual NMT capable of leveraging contextual signals such as film metadata and document-level information to improve translation quality, as well as enabling better control of phenomena such as speaker's gender and formality register. The mechanism for delivering context in the model involves converting the context fields into equal-sized vectors via sentence embedding. The resulting vector sequence is inputted into a distinct Transformer encoder. Additionally, we employ the context specificity evaluation method outlined in \newcite{vincent2024referenceless}, which relies on the pointwise mutual information (PMI). In this method, PMI quantifies the degree of co-occurrence between tokens in a translation hypothesis and the respective context.

\section{Experimental Setup} \label{eamt24:es}
The primary objective of our case study was to investigate whether post-editing MT is a cost-effective alternative to FST in our workflow, and to what extent domain-adapted training data and the utilisation of context have an impact in this area. Guided by the availability of resources, we operated in two language pairs: \textsc{en-de} and \textsc{en-fr} and considered four versions of the text in each, including MT outputs from three systems:
\begin{enumerate}[noitemsep]
    \item \textsc{Google}\footnote{\url{https://translate.google.com/}}, a general-purpose NMT engine used in previous work.
    \item \textsc{Base-NMT}, a non-contextual Transformer-based translation model parameter-matched to \textsc{MTCue} and trained on the same data (except context).
    \item \textsc{MTCue} system~\cite{vincent-etal-2023-mtcue}, a multi-encoder Transformer.
\end{enumerate}
We also operated on the human translations of the test set (\textsc{Ref}) approved for production.\footnote{This baseline is omitted during automatic evaluation (in fact, it is used as the reference text to calculate the automatic metrics), but is used as a baseline in the human evaluation, where the professionals are asked to post-edit this already sufficiently good text.}. For both \textsc{MTCue} and \textsc{Base-NMT}, we trained the models after~\newcite{vincent2024referenceless}, \S 4.1, in the \textsc{Overlap} setting which mimics a scenario with access to prior episodes of a tested series for training (a sample is presented in Appendix F of that work). We operated on sentence-level translations, with \textsc{MTCue} using the context for each sentence in its dedicated space.
\subsection{Automatic evaluation}
We conducted a pre-emptive automatic evaluation to confirm the feasibility of the human evaluation study. We used BLEU~\cite{papineni-etal-2002-bleu} and \textsc{Comet}~\cite{rei-etal-2020-comet} as translation quality metrics. Additionally, to measure context specificity, we measured the PMI between contextual and non-contextual translations~\cite{vincent2024referenceless}. We compared the outputs of the machine translation systems (\textsc{Base-NMT}, \textsc{Google}, \textsc{MTCue}) against the reference (\textsc{Ref}). 

\subsection{Post-Editing Setup and Metrics}
\begin{figure*}[!ht]
    \centering
    \includegraphics[width=.9\textwidth]{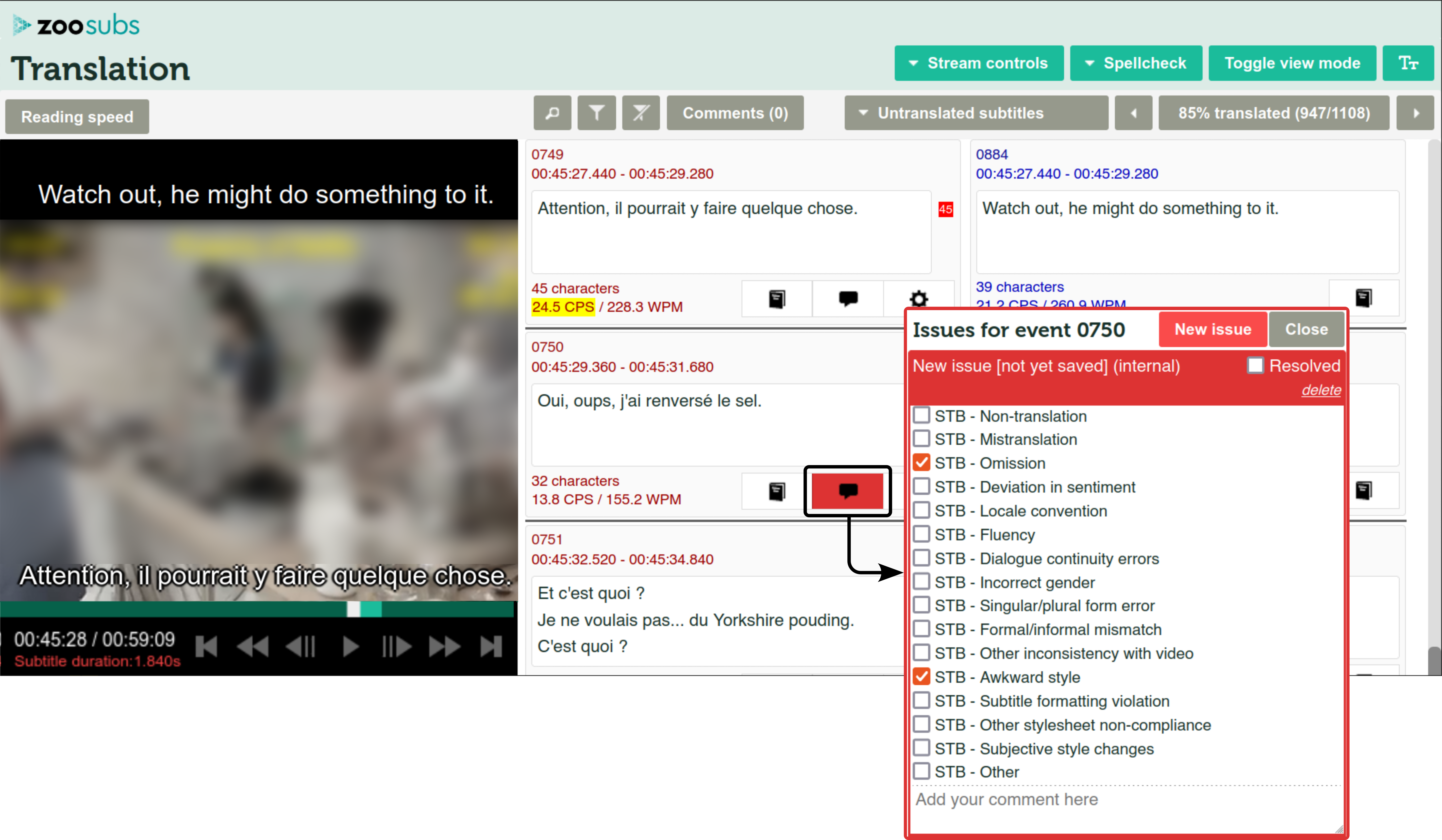}
    \caption{A compressed snapshot of \subs{}.}
    \label{fig:zoosubs}
\end{figure*}

The human evaluation aspect of the study is interpreted as the effort required to post-edit the translations to a production standard, and captured in the \textbf{number of errors, keystrokes} and \textbf{total edit time}. The task was performed by professional HTs and PEs using \subs{}, an in-house software application belonging to ZOO Digital, built to facilitate manual translation of video material (\autoref{fig:zoosubs}). The software's interface displays the video material along with timed subtitles in the original language. The \textit{target stream}, i.e. the set of text boxes provided to the right of the source stream, is where the HTs input their translations to the desired language. It can optionally be pre-populated with \enquote{draft} translations -- a setting we opted for in this study -- allowing post-editors to edit, divide or combine the segments as they see fit.

To make amendments to a segment, the PE needs to click on its box. From that point, the system tracks the time spent editing the box and the number of keystrokes made. These metrics are recorded for each box separately and taken into account only if the post-edited text differs from the original. After applying modifications, an \textbf{Issues for event} window appears for the user to specify the purpose of the changes by selecting errors from a predefined list, optionally providing text commentary. We leveraged this functionality of \subs{} to measure the total and average time and number of keystrokes made by HTs and PEs given some pre-existing translations. We also measured the number of selected errors. For this project, we created a bespoke taxonomy of errors (\autoref{he:errors}) based on translation errors reported in previous work~\cite{freitag-etal-2021-experts,sharou-specia-2022-taxonomy}, the original list of issues already present in the \subs{} system and relevant errors from previous work (\S \ref{eamt24:rw}). Error categories from the aforementioned sources were compiled together and curated to fit the study requirements\footnote{We uploaded a draft taxonomy to \subs{}, and the first author performed a test evaluation against a stream with $446$ segments to validate the list. As a result, some errors were split into more granular categories, some were renamed and some generalised.} 

\begin{table*}[ht!]
\centering
\resizebox{\textwidth}{!}{
\begin{tabular}{@{}rp{14.5cm}@{}}
\toprule
Type & Description \\ \midrule
\textbf{Translation quality} & \\
\textit{Catastrophic translation} & Impossible to post-edit, must be translated from scratch. \\
\textit{Mistranslation} & Incorrect. Does not preserve the meaning or function of the source. \\
\textit{Omission} & Part of the source text was left untranslated. \\
\textit{Deviation in sentiment} & Does not preserve the sentiment of the source (e.g. does not match the expressed excitement), or negates the sentiment (e.g. from positive to negative). \\
\textit{Locale convention} & Violates locale convention, e.g. currency and date format. \\
\textit{Fluency} & Contains punctuation, spelling and grammar errors. \\ \midrule
\textbf{Context} & \\
\textit{Incorrect gender} & Misgenders the speaker or the addressed person(s). \\
\textit{Incorrect plurality} & Incorrectly refers to a single person when a group is addressed, or vice versa. \\
\textit{Wrong formality} & Expressed in informal style or uses informal addressing when should use formal, or vice versa. \\ 
\textit{Other inconsistency with video} & Contains inconsistencies with the video material not falling within any of the above. \\\midrule
\textbf{Style} & \\
\textit{Subtitle formatting violation} & Violation of the subtitle blocking guidelines. \\
\textit{Other style sheet non-compliance} & Does not conform to the provided style sheet. \\
\textit{Awkward style} & The style of the translation does not reflect the style of the source sentence and/or the context. \\
\textit{Subjective style changes} & The translation is acceptable but the editor suggests improvements in style. \\ \midrule
\textbf{Other} & Error of type not found above (use text box provided). \\ \bottomrule
\end{tabular}}
\caption{List of errors provided to the human evaluators during the campaign.}
\label{he:errors}
\end{table*}

\paragraph{Worker setup}
The PEs operated on seven episodes from three TV series of varying genres: a fictional series about space exploration, a documentary exploring aspects of everyday life, and a family cooking competition show. They were unaware that some of the text they worked with was machine translated, but were told that it was for a research project and asked to relax some constraints such as adhering to the reading speed limits. In addition, we asked four HTs (two to German, two to French) to translate one episode of the cooking show from scratch in \subs{} so we could compare their effort to that of post-editors.
\begin{table}[ht!]
\resizebox{\linewidth}{!}{
\begin{tabular}{@{}rcc|cc|ccc@{}}
\toprule
\textbf{Series} 
& \multicolumn{2}{c|}{\textbf{\textsc{A}}}
& \multicolumn{2}{c|}{\textbf{\textsc{B}}}
& \multicolumn{3}{c}{\textbf{\textsc{C}}} \\ \midrule
\textbf{Ep. ID} & A1 & A2 & B1 & B2 & C1 & C2 & C3 \\ \midrule
\textbf{PE.1} & \Reff\textit{\textsc{Ref}} & \MTQ\textit{\textsc{MTCue}} & \Goog\textit{\textsc{Google}} & \BaseN\textit{\textsc{Base-NMT}} & \Reff\textit{\textsc{Ref}} & \MTQ\textit{\textsc{MTCue}} & \Goog\textit{\textsc{Google}} \\
\textbf{PE.2} & \BaseN\textit{\textsc{Base-NMT}} & \Reff\textit{\textsc{Ref}} & \MTQ\textit{\textsc{MTCue}} & \Goog\textit{\textsc{Google}} & \BaseN\textit{\textsc{Base-NMT}} & \Reff\textit{\textsc{Ref}} & \MTQ\textit{\textsc{MTCue}} \\
\textbf{PE.3} & \Goog\textit{\textsc{Google}} & \BaseN\textit{\textsc{Base-NMT}} & \Reff\textit{\textsc{Ref}} & \MTQ\textit{\textsc{MTCue}} & \Goog\textit{\textsc{Google}} & \BaseN\textit{\textsc{Base-NMT}} & \Reff\textit{\textsc{Ref}} \\
\textbf{PE.4} & \MTQ\textit{\textsc{MTCue}} & \Goog\textit{\textsc{Google}} & \BaseN\textit{\textsc{Base-NMT}} & \Reff\textit{\textsc{Ref}} & \MTQ\textit{\textsc{MTCue}} & \Goog\textit{\textsc{Google}} & \BaseN\textit{\textsc{Base-NMT}} \\ \midrule
\textbf{HT.1} & \cellcolor{White} \textit{From Scratch} & \cellcolor{gray} & \cellcolor{gray} & \cellcolor{gray} & \cellcolor{gray} & \cellcolor{gray} & \cellcolor{gray} \\
\textbf{HT.2} & \cellcolor{White} \textit{From Scratch} & \cellcolor{gray} & \cellcolor{gray} & \cellcolor{gray} & \cellcolor{gray} & \cellcolor{gray} & \cellcolor{gray} \\ \bottomrule
\end{tabular}}
\caption{Work assignment to PEs and HTs in the human evaluation campaign used for both language pairs.}
\label{eamt24:pe-setup}
\end{table}
For each of the seven episodes, the PEs were asked to post-edit one out of four versions of the text, corresponding to the list outlined in \S \ref{eamt24:es}. We included the human references (\textsc{Ref}) to account for the fact that PEs can sometimes post-edit a translation even when the original one is valid. Our setup ensured that the same PE evaluated the output for each episode exactly once (i.e. does not see two different versions of the same text) (\autoref{eamt24:pe-setup}). When referring to individual PEs, we use the notation \textbf{PE.[L][i]}, where \textbf{L} $\in$ \{\textbf{G} (German), \textbf{F} (French)\}, and \textbf{i} denotes the PE ID $\in [1,4]$.

\paragraph{Details regarding the PEs} \label{eamt24:pe-details} 
The recruited PEs and HTs were professionals within the subtitle domain and freelance employees of \textsc{ZOO Digital}. They were informed that the undertaken work was carried out for a research project, but nevertheless, they were paid for their effort at competitive PE and HT rates, standard within the company for this type of work. Information about the PEs' and HTs' years of experience (YOE) was collected to shed more light on the findings (\autoref{tab:pe-basic-info}). They also answered a short survey about their views regarding machine translation, discussed in detail in \S \ref{eamt24:he-pe-survey}:

\begin{enumerate}[nosep]
    \item Which one would you prefer: translating a stream from scratch or completing a quality check on (post-editing) a stream? Why?
    \item What are your views on the use of machine translation in the industry?
    \item In your view, are there benefits to post-editing translations over translating from scratch?
\end{enumerate}

\begin{table}[ht!]
\centering
\resizebox{\linewidth}{!}{
\begin{tabular}{@{}r|cccc|cccc@{}}
\toprule
 & \multicolumn{4}{c}{English-to-French} & \multicolumn{4}{c}{English-to-German} \\ \midrule
 & \textbf{PE.F1} & \textbf{PE.F2} & \textbf{PE.F3} & \textbf{PE.F4} & \textbf{PE.G1} & \textbf{PE.G2} & \textbf{PE.G3} & \textbf{PE.G4} \\ \midrule
Translation YOE & 15 & 8 & 3 & 20 & 7 & 18 & 8 & 17 \\
YOE in subtitles & 8 & 6 & 1.5 & 20 & 7 & 5 & 8 & 7 \\
YOE in post-editing & 8 & 6 & 3 & 10 & 5 & 5 & 1 & 3 \\
Post-editing training? & \cmark & \cmark & \cmark & \cmark & \xmark & \xmark & \xmark & \xmark \\
Prefer post-editing? & \cmark & \cmark & \xmark & \cmark & \cmark/\xmark & \xmark & \xmark & \xmark \\ \bottomrule
\end{tabular}}
\caption{Details regarding employed PEs.}
\label{tab:pe-basic-info}
\end{table}
All French HTs had training in post-editing, and three out of four preferred it to translating from scratch, while no German HTs had received such training in the past, and all but one strictly preferred FST. All PEs had at least one YOE in post-editing and one and a half in the subtitle domain. Although the HTs within both pairs had a similar amount of experience in translation in general and in the subtitle domain ($11.5 \pm 6.5$ for French vs $12.5 \pm 5.0$ for German), the French HTs had the advantage in terms of YOE in both subtitling (a mean difference of $2.1$ YOE) and post-editing (a mean difference of $3.3$ YOE).

\section{Results of Automatic Evaluation} \label{eamt24:ae}
\begin{figure*}[h!]
\includegraphics[width=\textwidth]{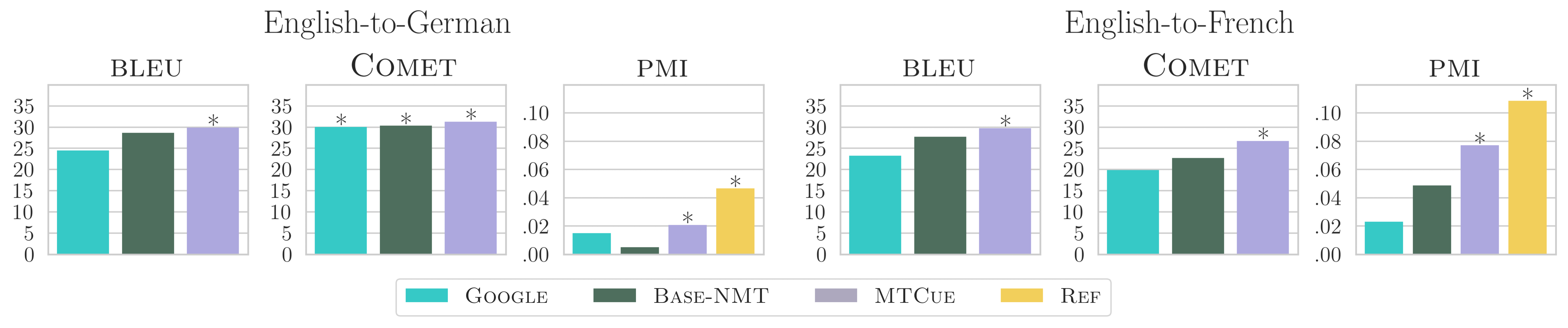}
    \caption{BLEU, \textsc{Comet} and PMI scores obtained by the evaluated models. Asterisks (*) over bars indicate the best result along with all statistically indistinguishable results computed either via bootstrap resampling (or t-test for PMI), $p=0.05$.}
    \label{fig:eamt24-eval}
\end{figure*}

The automatic evaluation results (\autoref{fig:eamt24-eval}) suggest that \textsc{MTCue} was the best-performing system and \textsc{Google} the worst-performing for both language pairs. Interestingly, for \textsc{en-de}, the BLEU and \textsc{Comet} score differences varied in magnitude, to the point of \textsc{Comet} judging all three systems as on par. A possible cause was the discrepancy in hypothesis length (the reference text uses $7.04$ words per segment, \textsc{Base-NMT}: $7.06$, \textsc{MTCue}: $7.06$, \textsc{Google}: $8.29$). Since \textsc{Comet}'s calculation involves comparing sentence embeddings of the hypothesis and the reference, including more words or phrases in the hypothesis may lead to a closer similarity match, inflating the score even if the additional tokens are redundant or even harmful to quality. BLEU does not have this problem as it is based on string matching~\cite{papineni-etal-2002-bleu}. As per the PMI scores, the professional translations (\textsc{Ref}) consistently exhibited the highest context specificity. However, \textsc{MTCue} was on par with this reference score in both cases and was consistently better than the other two systems. \textsc{MTCue} therefore shows promise at addressing the context-related issues in subtitle translation.

\section{Results of the Post-Editing Study} \label{eamt24:he}
This section analyses the results of the post-editing study: the translation errors (\S \ref{eamt24:he-error}), the post-editing effort (\S \ref{eamt24:he-effort}), and finally, the post-campaign survey responses (\S \ref{eamt24:he-pe-survey}).

Due to the unprecedented nature of this work in the company, the professionals' contract allowed them to withdraw if they found the compensation insufficient for the requested work. At the midpoint of the campaign, two PEs (\textbf{PE.G1} and \textbf{PE.G3}) contacted the project manager to express concerns regarding the quality of the MT outputs, asserting that the task potentially required more effort than FST. To compromise, they proposed narrowing the scope of the remaining work to error identification and marking, without making the necessary corrections. This meant we would not obtain the effort metrics for the two PEs. Consequently, while the error analysis in \S \ref{eamt24:he-error} includes both language pairs, the effort analysis in \S \ref{eamt24:he-effort} does not include results from \textbf{PE.G1} or \textbf{PE.G3}.

\subsection{Error Analysis} \label{eamt24:he-error}
An initial inspection of the results indicated that each PE marked a significantly different total number of errors (e.g. \textbf{PE.F1} marked $232$ errors total while \textbf{PE.F4} marked $878$). This made direct comparison of the error counts across systems unreliable as each PE also post-edited a different number of segments for each system (cf. \autoref{eamt24:pe-setup}). With seven episodes and four different versions of the text, for each PE there is a version of text they would only have seen one episode from. For example, in \autoref{eamt24:pe-setup}, \textbf{PE.1} is assigned two episodes for \textsc{Ref}, \textsc{MTCue} and \textsc{Google}, but only one for \textsc{Base-NMT}. In this example, if \textbf{PE.1} generally marked fewer errors than others, \textsc{Base-NMT} would be disproportionately rewarded. 

To make the measurements comparable, we normalised them by computing a \textit{normalisation coefficient} $h$ for each PE and then multiplying their error counts for each category by their $h$. Let $\textsc{Err}_{PE_i,c}$ denote the number of errors within the category $c$ for the $i$-th PE. We compute the normalised count $\widehat{\textsc{Err}}_{PE_i,c}$ as described by \autoref{eq:norm-err}.
\begin{equation} \label{eq:norm-err}
    \begin{split}
        \widehat{\textsc{Err}}_{PE_i,c} &= \textsc{Err}_{PE_i,c} \times h_i \\
        \text{where }h_i &= \frac{
            \max(
                \textsc{Err}_{PE_j,total}; j \in \{1,4\}
                )
        }{\textsc{Err}_{PE_i,total}} \\    
    \end{split}
\end{equation}

We report the total error counts as well as the normalisation multipliers in \autoref{eamt24:norms}.

\begin{table}[h!]
\centering
\resizebox{\linewidth}{!}{
\begin{tabular}{@{}rcc|rcc@{}}
\toprule
\multicolumn{3}{c}{\textbf{English-to-German}} & \multicolumn{3}{|c}{\textbf{English-to-French}} \\ \midrule
\textit{PE ID} & \textit{Error count} & $h$ & \textit{PE ID} & \textit{Error count} & $h$ \\
\textbf{PE.G1} & $1526$ & $1.76$ & \textbf{PE.F1} & $232$ & $14.68$ \\
\textbf{PE.G2} & $2452$ & $1.10$ & \textbf{PE.F2} & $182$ & $18.71$ \\
\textbf{PE.G3} & $2690$ & $1.0$ & \textbf{PE.F3} & $3406$ & $1.0$ \\
\textbf{PE.G4} & $1832$ & $1.47$ & \textbf{PE.F4} & $878$ & $3.88$ \\ \bottomrule
\end{tabular}}
\caption{Error counts and values of $h$ for each PE.}
\label{eamt24:norms}
\end{table}

\paragraph{Error post-processing} 
To facilitate post-editing in \subs{}, MT outputs had to be adapted to match the subtitle format. Quality checks of translations conducted in \subs{} normally require the users not just to ensure the correctness of translations but also that the subtitles comply with strict guidelines\footnote{This includes adhering to reading speed and length limits, balancing the length of the top and bottom subtitle, disambiguation of speaker turns with colours or dashes, and applying appropriate formatting, as specified by a style sheet.}. Typical MT systems, like the ones used in this project, are not designed to create translations conforming to these stringent guidelines, and the primary goal of this study was to identify the impact of the translation errors alone. To faithfully replicate the normal work environment of the PEs, we applied a greedy reformatting tool (built into \subs{}) to reformat our translations as subtitles. We made it clear that the project is centred on the correctness of translations, not the subtitle formatting. Still, to ensure that the translation and non-translation errors are kept separate, we included two environment-specific errors for the workers to select from: \textit{Subtitle formatting violation} covering cases where the subtitle is not split to optimally adhere to segmentation guidelines; and \textit{Other style sheet non-compliance} where a rule outlined in the style sheet from the client company was not followed, such as custom punctuation conventions.

\begin{table}[h!]
    \resizebox{\linewidth}{!}{
    \begin{tabular}{rp{0.7\linewidth}}
        \textbf{Example 1} & \hfill Target: German \\ \midrule
        Source & Can I take a look at what you're doing by any chance? \\
        \textsc{Base-NMT} (\xmark) & Kann ich mir \brick{zufällig} ansehen, was \brick{du} \brick{[BR]} \brick{machst}? \\
        Post-ed. & Kann ich mir \olive{vielleicht} ansehen, \olive{[BR]} was \olive{Sie da machen}? \\
        Errors & \textit{Mistranslation} \\
        & \textit{Subtitle formatting violation} \\
        & \textit{Formal/informal mismatch} \\
    \end{tabular}}
    \notag
\end{table}
\label{eamt24:example-0}

In some instances, a PE would encounter both translation and non-translation errors within the same segment, as exemplified in \hyperref[eamt24:example-0]{\textbf{Example 1}}, where both translation errors (\textit{Mistranslation} of \textit{by any chance} and \textit{Formal/informal mismatch} of \textit{you're doing}) and non-translation errors (\textit{Subtitle formatting violation} of the position of the subtitle break) are present. In such cases, we (i) disregard the non-translation error counts, and (ii) correct the effort rates (editing time and keystrokes) to account solely for translation-related errors. To precisely gauge the latter, we employed a correction method: let $\textsc{Err}_{non-translation}$ and $\textsc{Err}_{translation}$ be the total effort expended by a PE on a segment that had only non-translation and only translation errors marked, respectively. We calculated translation share (TS) as follows:
\[ \text{TS} = \frac{\textsc{Err}_{translation}}{\textsc{Err}_{translation} + \textsc{Err}_{non-translation}} \]
We then used it to calculate the estimated share of the effort spent on translation in segments that had both errors marked by multiplying TS by the total effort spent on a segment with both error types.\footnote{For example, if a PE took three seconds for translation errors and two seconds for non-translation errors on average, where they marked both types we multiplied their total effort for that segment by $\frac{3}{3+2}$.}

Finally, since the \textbf{Other} category was used substantially, we parsed the contents of the optional description text box. The most commonly reported \textbf{Other} errors were \enquote{Grammar}, \enquote{Punctuation}, \enquote{Timing}, \enquote{SGP} (spelling, grammar, punctuation) and \enquote{Literal translation}. Such errors ($69.3\%$) were removed from the \textbf{Other} category and pigeonholed as appropriate (e.g. \enquote{Grammar} as \textit{Fluency}). More complex comments such as “wissen Sie should not be in the translation” were left categorised as \textit{Other} ($30.7\%$).

\begin{table*}[htb!]
\centering
\scalebox{0.8}{
\begin{tabular}{rr|ccc|c}
\toprule
& & \multicolumn{4}{c}{Normalised count} \\
& Error type & \textsc{Google} & \textsc{Base-NMT} & \multicolumn{1}{c}{\textsc{MTCue}} & \textsc{Ref} \\ \midrule
\parbox[t]{2mm}{\multirow{18}{*}{\rotatebox[origin=c]{90}{\textbf{English-to-German}}}} 

& \textbf{Translation quality} & $13.12 \pm 14.46$ & $\underline{8.70 \pm 11.67}$ & $\mathbf{8.49 \pm 10.90}$ & $4.56 \pm 5.14$ \vspace{.5em}\\
& \textit{Catastrophic translation} & $\underline{0.50 \pm 0.27}$ & $\mathbf{0.46 \pm 0.18}$ & $\underline{0.88 \pm 0.95}$ & $0.72 \pm 0.68$ \\
& \textit{Mistranslation} & $\underline{26.99 \pm 8.58}$ & $\mathbf{25.69 \pm 7.67}$ & $\underline{26.74 \pm 6.15}$ & $8.76 \pm 5.51$ \\
& \textit{Omission} & $\mathbf{0.26 \pm 0.15}$ & $2.32 \pm 2.20$ & $3.54 \pm 2.79$ & $5.38 \pm 6.75$ \\
& \textit{Deviation in sentiment} & $\underline{1.11 \pm 0.66}$ & $\mathbf{0.83 \pm 0.30}$ & $\underline{1.25 \pm 0.88}$ & $5.23 \pm 4.40$ \\
& \textit{Locale convention} & $2.04 \pm 0.00$ & $\underline{0.94 \pm 0.46}$ & $\mathbf{0.61 \pm 0.30}$ & $0.91 \pm 1.03$ \\
& \textit{Fluency} & $16.88 \pm 15.22$ & $\underline{9.54 \pm 11.17}$ & $\mathbf{7.10 \pm 6.52}$ & $4.18 \pm 3.65$ \\ \cmidrule{2-6}

& \textbf{Context} & $5.34 \pm 5.68$ & $\underline{2.64 \pm 3.45}$ & $\mathbf{2.21 \pm 2.55}$ & $1.18 \pm 1.13$ \vspace{.5em}\\
& \textit{Incorrect gender} & $\underline{2.20 \pm 1.58}$ & $\underline{1.69 \pm 1.90}$ & $\mathbf{1.43 \pm 1.17}$ & $1.60 \pm 1.19$ \\ 
& \textit{Plural/singular form error} & $\underline{0.99 \pm 0.81}$ & $\mathbf{0.80 \pm 0.63}$ & $\underline{1.19 \pm 1.24}$ & $0.33 \pm 0.00$ \\ 
& \textit{Formal/informal mismatch} & $11.31 \pm 4.55$ & $\underline{5.29 \pm 4.60}$ & $\mathbf{3.86 \pm 3.60}$ & $1.19 \pm 1.31$ \\ \cmidrule{2-6}

& \textbf{Style} & $\underline{12.19 \pm 9.79}$ & $\mathbf{8.12 \pm 6.59}$ & $\underline{9.88 \pm 7.83}$ & $3.77 \pm 3.86$ \vspace{.5em}\\
& \textit{Awkward style} & $17.70 \pm 7.76$ & $\mathbf{11.82 \pm 5.21}$ & $\underline{13.11 \pm 7.04}$ & $4.70 \pm 4.34$ \\
& \textit{Subjective style changes} & $\underline{2.55 \pm 2.09}$ & $\mathbf{1.65 \pm 1.59}$ & $\underline{2.33 \pm 2.28}$ & $2.13 \pm 2.52$ \\ \cmidrule{2-6}

& \textbf{Other} & $\underline{2.12 \pm 3.43}$ & $\underline{3.26 \pm 4.48}$ & $\mathbf{2.10 \pm 2.46}$ & $3.39 \pm 5.88$ \\ \midrule 
& \textbf{Total} & $9.58 \pm 11.35$ & $\underline{6.44 \pm 9.05}$ & $\mathbf{6.41 \pm 8.82}$ & $3.86 \pm 4.70$ \\ \midrule \midrule

\parbox[t]{2mm}{\multirow{18}{*}{\rotatebox[origin=c]{90}{\textbf{English-to-French}}}} 

& \textbf{Translation quality} & $20.01 \pm 23.05$ & $\mathbf{9.27 \pm 9.52}$ & $\underline{10.21 \pm 8.88}$ & $6.60 \pm 5.08$ \vspace{.5em} \\
& \textit{Catastrophic translation} & $\underline{3.41 \pm 1.38}$ & $\mathbf{2.25 \pm 2.39}$ & $\underline{2.86 \pm 3.03}$ & $2.51 \pm 3.26$ \\
& \textit{Mistranslation} & $38.80 \pm 14.35$ & $\underline{22.73 \pm 8.49}$ & $\mathbf{20.10 \pm 7.34}$ & $7.24 \pm 3.61$ \\
& \textit{Omission} & $\mathbf{2.40 \pm 2.40}$ & $\underline{3.91 \pm 1.49}$ & $5.56 \pm 4.09$ & $7.48 \pm 5.13$ \\
& \textit{Deviation in sentiment} & $\mathbf{5.93 \pm 5.90}$ & $\underline{7.82 \pm 6.09}$ & $11.59 \pm 0.00$ & $6.74 \pm 3.03$ \\
& \textit{Locale convention} & $4.29 \pm 2.49$ & $0.73 \pm 0.51$ & $\mathbf{0.21 \pm 0.00}$ & $0.63 \pm 0.00$ \\
& \textit{Fluency} & $30.83 \pm 31.77$ & $\underline{7.28 \pm 3.75}$ & $\mathbf{5.92 \pm 4.18}$ & $7.82 \pm 7.35$ \\ \cmidrule{2-6}

& \textbf{Context} & \underline{$5.41 \pm 3.64$} & $6.09 \pm 4.26$ & $\mathbf{3.86 \pm 3.11}$ & $1.29 \pm 1.07$ \vspace{.5em}\\
& \textit{Incorrect gender} & $\mathbf{3.49 \pm 2.59}$ & $6.96 \pm 5.57$ & $\underline{4.77 \pm 3.98}$ & $0.49 \pm 0.44$ \\ 
& \textit{Plural/singular form error} & $4.50 \pm 1.92$ & $5.84 \pm 4.60$ & $\mathbf{1.97 \pm 0.62}$ & $0.00 \pm 0.00$ \\ 
& \textit{Formal/informal mismatch} & $\underline{7.44 \pm 4.63}$ & $\underline{5.58 \pm 3.76}$ & $\mathbf{4.23 \pm 2.93}$ & $1.69 \pm 1.10$ \\ \cmidrule{2-6}

& \textbf{Style} & $11.05 \pm 7.07$ & $10.35 \pm 3.69$ & $\mathbf{3.41 \pm 2.53}$ & $5.55 \pm 3.41$ \vspace{.5em}\\
& \textit{Awkward style} & $11.13 \pm 7.46$ & $9.55 \pm 1.27$ & $\mathbf{2.89 \pm 2.76}$ & $4.10 \pm 1.28$ \\
& \textit{Subjective style changes} & $\underline{10.94 \pm 8.16}$ & $11.15 \pm 5.52$ & $\mathbf{4.18 \pm 2.87}$ & $6.28 \pm 4.09$ \\ \cmidrule{2-6}

& \textbf{Other} & $37.20 \pm 52.68$ & $\mathbf{11.19 \pm 16.44}$ & $\underline{23.67 \pm 29.23}$ & $27.05 \pm 24.68$ \\ \midrule
& \textbf{Total} & $17.02 \pm 25.78$ & $\mathbf{8.84 \pm 9.20}$ & $\underline{9.63 \pm 13.85}$ & $8.83 \pm 12.84$ \\ \bottomrule
\end{tabular}
}
\caption{Counts of errors flagged by the PEs for each system. Excluding \textsc{Ref}, the best result in each row is highlighted and all statistically indistinguishable results are underlined (one-tailed t-test, confidence interval of $80\%$, $p=0.2$). Error rates for categories in bold (e.g. \textbf{Style}) are calculated based on all errors within the category.}
\label{tab:error-results}
\end{table*}

\paragraph{Results}
The calculated normalised counts of errors within each category (\autoref{tab:error-results}) suggest that \textsc{MTCue} performs no worse than both non-contextual MT systems overall (row \textbf{Total}), while performing significantly better in the \textbf{Context} and \textbf{Style} categories in \textsc{en-fr}, pointing to gains related to the use of context information. 

The most frequently flagged errors in both language pairs were consistently \textit{Mistranslation} and \textit{Fluency}. \textit{Mistranslation} was reported a similar number of times for all three machine translation systems in \textsc{en-de} and three times less frequently for post-editing \textsc{Ref}. This gap was similar in \textsc{en-fr}, though within the MT systems themselves, the \textsc{Google} system had a significantly higher error rate for \textit{Mistranslation} errors ($38.80$ mean) than the next best system, i.e. \textsc{Base-NMT} ($22.73$); the contextual \textsc{MTCue} achieved an even lower rate of $20.10$. Interestingly, \textsc{MTCue} also produced outputs of higher \textit{Fluency} than other systems, even surpassing \textsc{Ref} for \textsc{en-fr}, though insignificantly at the selected confidence interval ($80\%$).

In both language pairs, the \textit{Omission} error was consistently marked the fewest times in \textsc{Google}-generated text (see \textbf{Translation quality} $\rightarrow$ \textit{Omission}). In both cases, \textsc{Ref} scored significantly above the mean. This is unsurprising: translations authored by the general-purpose \textsc{Google} engine tend to be overly literal and faithful to the source, while in the domain of dialogue, the HT often needs to let go of individual features of the source text or opt for alternative expressions to maintain the brevity and dynamics of the source dialogue, leading to spontaneous omissions in the reference translations. To exemplify, \textsc{Google} consistently unnecessarily translated the English \textit{\enquote{(...), you know,}} to \textit{\enquote{(...), wissen Sie,}} in German, necessitating additional post-editing in our study. A similar error was typically avoided by the other systems, due to their data-learned preference for brevity and dynamically expressive language. As a result, both systems were marked with \textit{Omission} more times than \textsc{Google}. In fact, \textsc{MTCue} scored even more \textit{Omission}s than \textsc{Base-NMT}, suggesting that \textsc{MTCue}'s omission behaviour more closely matches that of professional HTs. Other \textbf{Translation quality} errors were relatively infrequent and with insignificant differences between systems.

To capture context-related issues, we provided categories for the most frequent contextual errors: \textit{Incorrect gender}, \textit{Plural/singular form} and \textit{Formal/informal mismatch}.
Since the perception of speaking style in dialogue is subjective and difficult to gauge, we did not provide explicit ways for the PEs to mark speaker style errors to avoid biasing them towards thinking in terms of what is a characteristic way of expression for the given speaker. Instead, we provided loose categories for \textbf{Style}, with the intention of collecting measurements of how often the PEs feel the need to alter the style of the translations. Since all of the post-edited content is dialogue, the style of the translation can be directly associated with the style of the speaker's expression. Our findings regarding some \textbf{Context} categories (\textit{Incorrect gender}, \textit{Formal/informal mismatch}) are consistent between the two language pairs, and \textsc{MTCue} was found to be superior in most categories in both cases, with the overall score for the \textbf{Context} category being significant at $80\%$ confidence for \textsc{en-fr}. The \textit{Plural/singular} form error required few corrections in \textsc{en-de} (where \textsc{Base-NMT} was found superior to \textsc{MTCue}) and more in \textsc{en-fr} (where \textsc{MTCue} was found superior).

The findings from the \textbf{Style} category also work in favour of contextual MT, where it was found comparable to non-contextual systems for the \textsc{en-de} pair and significantly better than them for the \textsc{en-fr} pair, requiring the fewest style-based adjustments, even fewer than \textsc{Ref}. Within the \textsc{en-de} pair, \textit{Subjective style changes} were flagged only up to $4-5$ times per $100$ segments for any system, and a consistent number of times between systems, and \textit{Awkward style} was flagged the fewest times for \textsc{Ref} ($4.68$ on average), much less frequently than for the other systems, among which \textsc{Google} required the most edits and \textsc{Base-NMT} the fewest.

Overall, our error count analysis suggests that within the \textsc{en-fr} pair, \textsc{MTCue} has significantly reduced the number of errors marked for contextual and stylistic reasons compared to non-contextual systems, while not degrading overall translation quality. The findings within the \textsc{en-de} pair are too variable to yield definitive conclusions but entail no degradation of quality leading from the inclusion of context, a significant improvement for contextual phenomena compared to \textsc{Google}, and highlight that \textsc{MTCue} makes the fewest contextual errors overall.

\subsection{Analysis of Effort and Quality} \label{eamt24:he-effort}
This section delves into the analysis of per-PE effort spent post-editing or translating the outputs of each system. Based on the observation that some measurements of editing time and keystrokes were out of the distribution, we normalised these by first computing the $97.5$th percentile for the given language pair and task (translation or post-editing) and set all per-segment measurements to be capped at that percentile. Our obtained percentiles were: $37$ seconds and $69$ keystrokes for translation, and $45$ seconds and $54$ keystrokes for post-editing.

\paragraph{Effort per PE}
\begin{figure*}[h!]
    \centering
    \includegraphics[width=\textwidth]{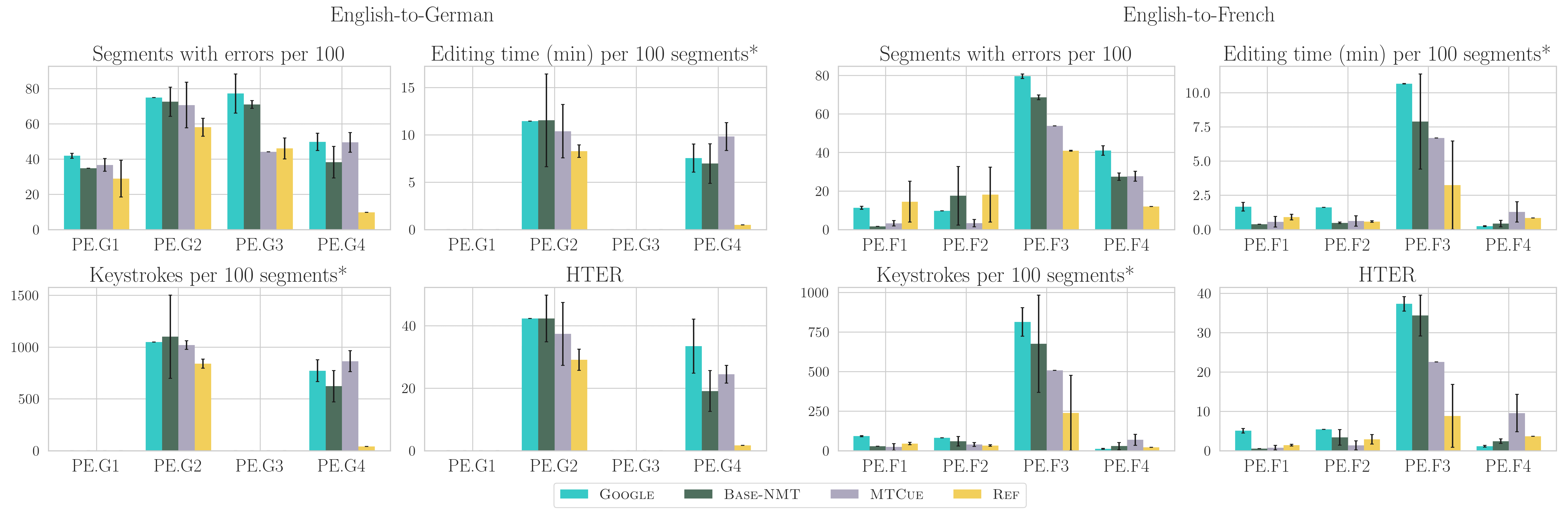}
    \caption{Effort for each PE within both language pairs.}
    \label{fig:effort-per-pe}
\end{figure*}

As per \autoref{fig:effort-per-pe}, the results for the \textsc{en-de} pair suggest that each PE contributed a similar effort. Interestingly, the error rate and effort measures of these PEs are closer in magnitude to the outlier \textbf{PE.F3} within the \textsc{en-fr} pair. Putting PEs from both pairs together we find an interesting correlation: those PEs who expressed a preference for post-editing marked significantly fewer errors overall. We suspect that professionals who expressed a preference for translation opted for spending any effort necessary to match the quality of the resulting text to what they would have produced from scratch, while the post-editing enthusiasts contributed fixed effort, possibly characteristic of their usual post-editing assignments.

The error rate for this pair points to \textsc{Google} as the system consistently requiring the most edits, and \textsc{Ref} the least, though only \textbf{PE.G4} made drastically fewer edits to this already production-ready text. Between \textsc{Base-NMT} and \textsc{MTCue}, \textbf{PE.G2} and \textbf{PE.G3} found \textsc{MTCue} to be less erroneous (and \textbf{PE.G3} found it to be on par with \textsc{Ref}), while \textbf{PE.G1} and \textbf{PE.G4} identified fewer errors in \textsc{Base-NMT}.

According to \textbf{PE.G2}, the quality of translations from \textsc{Google} and \textsc{Base-NMT} is comparable, requiring the most complex and laborious edits. \textsc{MTCue}'s hypotheses required less work from this PE, and \textsc{Ref} text still less. Results obtained from \textbf{PE.G4}'s edits are different, revealing next to no edits to the \textsc{Ref} text, (which could be interpreted as them being the least subjective of the PEs, only making edits when they are necessary). This PE found \textsc{MTCue} to require more edits than \textsc{Base-NMT} and on par with \textsc{Google}. Interestingly, even though editing \textsc{MTCue}'s outputs took more time and keystrokes, \textsc{Google}'s outputs yielded a HTER value about $10$ points higher than \textsc{MTCue}. Since \textsc{Google} is the more literal MT system, and \textsc{MTCue} produces more dialogue-like responses, these findings suggest that, other things being equal, a literal and overly long translation of dialogue may take less effort to post-edit than an incorrect platonic (dialogue-like) response, even if more profound edits are required.

\paragraph{Approach to \textsc{Ref}} 
Since the PEs were told about the research nature of the project, they might have approached this project with less vigilance than if the work was undertaken for actual clients. On the flip side, some may have eventually realised they were dealing with some MT outputs -- they were not told this explicitly -- and became more scrutinous as a result, expecting to make many more corrections than in a typical post-editing task. This would perhaps explain why some PEs took to post-editing \textsc{Ref} at rates sometimes matching the outputs of the MT systems, with three of them doing so at a rate of over $40$ errors per $100$ segments. 

\begin{figure}[ht!]
    \centering
    \includegraphics[width=\linewidth]{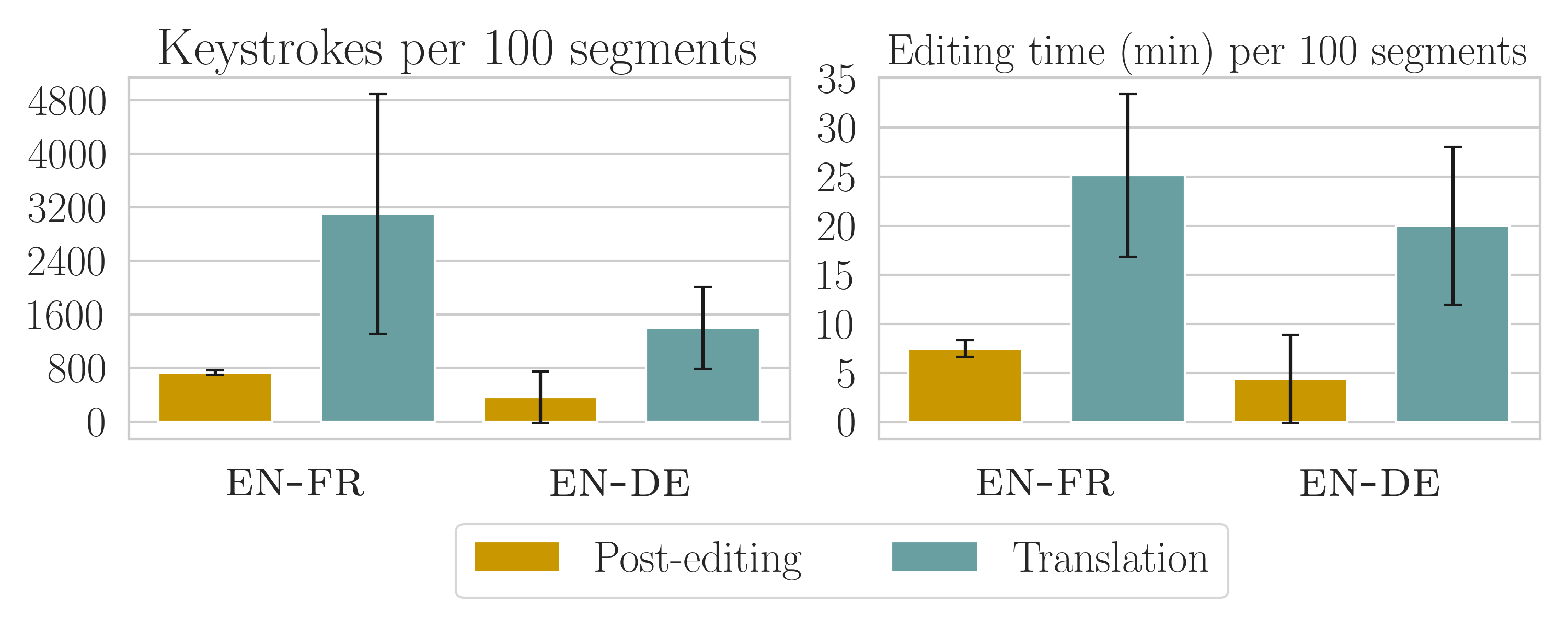}
    \caption{Effort comparison of FST and post-editing MT.}
    \label{fig:translation-effort}
\end{figure}

\paragraph{Comparison with translation effort}
In \autoref{fig:translation-effort} we compare the unnormalised post-editing effort (exclusive of \textsc{Ref}) to the FST effort for one episode of the cooking show. For both language pairs, FST required $4$ to $6$ times the effort of post-editing, by both measures.

\subsection{Analysis of the professionals' views on post-editing and MT} \label{eamt24:he-pe-survey}
Finally, we present the PEs' responses to a survey regarding views on post-editing and machine translation. Most of the German PEs expressed a preference for FST over post-editing, with three voicing frustration with MT's stiffness and literal nature, omitting aspects of the original text such as slang, gender agreement, references to the video and people's speaking styles. They view translation as a more creative process which can yield idiomatic and fluent translations. They also noted that post-editing currently demands more effort than translating from scratch at times, yet it is compensated at a lower rate than translation. To one PE, post-editing felt like damage control.

Conversely, three out of four French PEs expressed a preference for post-editing, justifying the choice with their specialisation. The fourth PE was dissatisfied with the amount of subtitle formatting errors within our project, commenting that FST would have focused more on content.

PEs in both languages agreed that MT can be a helpful tool, and praised the recent developments, but still concurred that the substantial gap in quality persists, and renders MT insufficiently competent to replace FST. However, they were optimistic about future developments within MT. The majority of PEs recognized the advantages of post-editing, such as the reduction of temporal effort in some cases and the potential to improve consistency in translating terminology, and enabling greater attention to detail. However, presently these benefits can fail to materialise in practice, emphasising the importance of further work on implementation quality of post-editing workflows.

\subsection{Examples of challenges} \label{eamt24:examples}
We present two examples of corrections made in the post-editing process to reflect what kind of corrections required attention as well as what mistakes need to be improved upon in the future.

\begin{table}[h!]
    \resizebox{\linewidth}{!}{
    \centering
    \begin{tabular}{rp{0.7\linewidth}}
        \textbf{Example 2} & \hfill Target: German \\ \midrule
        Source & No way, no way. \\
        Video context & \textit{The victorious family is in disbelief about their triumph.} \\
        \textsc{MTCue} (\xmark) & \brick{Auf keinen Fall.} \\
        & (`\textit{Under no circumstance.}') \\
        Post-ed. & \olive{Unmöglich.} \\
        & (`\textit{Unbelievable.}') \\
        Error & \textbf{Other}: \textit{inconsistency with video} \\
    \end{tabular}}
    \notag
\end{table}
\label{eamt24:example-4}

\hyperref[eamt24:example-4]{\textbf{Example 2}} presents a scenario where \textsc{MTCue} incorrectly interprets the exclamation \textit{No way} as \textit{Under no circumstance}, which fails to account for the sense of disbelief and amazement that the victorious family is experiencing. Such an interpretation relies strongly on the visual context, of which effective incorporation into the machine translation process in a multi-modal framework is an area for future work.

\begin{table}[h!]
    \centering
    \resizebox{\linewidth}{!}{
    \begin{tabular}{rp{0.7\linewidth}}
        \textbf{Example 3} & \hfill Target: German \\ \midrule
        Video context & \textit{Two cooks and a chopping board.} \\
        Source N & Get that Welly on that board. \\
        Reference N & Leg das Welly auf das Brett. \\
        \textsc{MTCue} (\xmark) & \brick{Stell} \brick{die} Welly auf das Brett. \\
        Post-ed. & \olive{Legt} \olive{das} Wellington auf das Brett. \\
        Error & \textit{Awkward style} \\
        \midrule
        Source N+1 & She's on. \\
        Reference N+1 & Es ist drauf. \\
        \textsc{MTCue} (\xmark) & \brick{Sie} ist \brick{dran.} \\
        Post-ed. & \olive{Ist drauf.} \\
        Error & \textbf{Other}: \textit{inconsistency with video} \\
    \end{tabular}}
    \notag
\end{table}
\label{eamt24:example-5}

\hyperref[eamt24:example-5]{\textbf{Example 3}} presents a two-error scenario. Firstly, \textsc{MTCue} uses the incorrect German preposition \textit{an}/\textit{dran} to translate the English \textit{on}, instead of the correct \textit{auf}/\textit{drauf} (\textit{on that board} $=$ \textit{auf das Brett}). The more interesting error comes from mistranslating \textit{She} as \textit{Sie}. The pronoun is a reference to pork Wellington, abbreviated to \textit{Welly} by the speaker, and incorrectly assigned the feminine article \textit{sie}, instead of the neuter \textit{das}. The speaker personifying the pork in Source N+1 (referring to it as \textit{She}) complicates things, and so even a document-level system could have trouble interpreting what \textit{Welly} actually is. The correct interpretation is crucial to selecting the right verb \textit{legen} over \textit{stellen} which should be used to translate \textit{get} when referring to meat. Though it was marked with an \textit{inconsistency with video} error, it is challenging to outline the minimal set of context information sufficient for the correct treatment of this example. The context of cooking, the light-hearted, casual character of the show and the manner of British speech, as well as what meal is being made and what the cooks are doing at the moment, all could aid this process. An important challenge for future contextual systems is going to be to discern which type of information is necessary and when.

\section{Conclusions and Future Work} \label{eamt24:c}
We have presented a case study on post-editing MT of subtitles for TV series in a multi-modal scenario, with a focus on contextual MT. We found that the MT models custom-trained on dialogue required less post-editing effort than the one-size-fits-all Google Translate, potentially due to the overbearing literalness and stiffness of the latter system's outputs. We also found that some post-editors amended production-approved human translations at high rates, with hypervigilance about dealing with MT as a possible cause. Our results did not determine a significant difference in post-editing effort between \textsc{MTCue} and \textsc{Base-NMT}. However, the inclusion of context in \textsc{MTCue} yielded fewer errors in the \textbf{Style}, \textbf{Context} and \textit{Fluency} categories, motivating our future exploration of context-inclusive models. We further found that post-editing any MT output required four to six times less technical and temporal effort compared to FST, making it a promising cost-effective venture. However, cognitive effort should be measured in future studies, given the exit survey sentiment that post-editing was sometimes harder and less interesting than FST. Our future experiments will employ larger cohorts of PEs and split them into groups who post-edit non-contextual and contextual inputs exclusively, so that clearer feedback can be collected, as well as to minimise the variance in effort. 

\section{Acknowledgements}
This work was completed as part of Sebastian Vincent's PhD, which was partially funded by ZOO Digital. Sebastian was also supported by the Centre for Doctoral Training in Speech and Language Technologies (SLT) and their Applications funded by UK Research and Innovation (grant number EP/S023062/1).

\bibliography{bib/custom,bib/references,bib/anthology}
\bibliographystyle{eamt24}
\end{document}